\pdfoutput=1
\documentclass[runningheads]{llncs}
\usepackage[version=3]{mhchem}
\usepackage{graphicx}
\begin{document}
\title{Geometric learning of the conformational dynamics of molecules using dynamic graph neural networks\thanks{This research was conducted under the Laboratory Directed Research and Development Program at Pacific Northwest National Laboratory, a multi-program national laboratory operated by Battelle for the U.S. Department of Energy.}}
\titlerunning{Molecular Dynamic Graph Neural Networks}
%
\author{Michael Hunter Ashby\orcidID{0000-0003-4468-649X}\newline \and
Jenna A. Bilbrey\orcidID{0000-0002-4199-7244}}
\authorrunning{M.H. Ashby and J.A. Bilbrey}
%
\institute{Pacific Northwest National Laboratory, Richland, WA, USA \\ 
\email{michael.ashby@pnnl.gov} \newline
\email{jenna.pope@pnnl.gov}}
\maketitle              
\begin{abstract}

We apply a temporal edge prediction model for weighted dynamic graphs to predict time-dependent changes in molecular structure. Each molecule is represented as a complete graph in which each atom is a vertex and all vertex pairs are connected by an edge weighted by the Euclidean distance between atom pairs. We ingest a sequence of complete molecular graphs into a dynamic graph neural network (GNN) to predict the graph at the next time step. Our dynamic GNN predicts atom-to-atom distances with a mean absolute error of 0.017 \AA, which is considered ``chemically accurate'' for molecular simulations. We also explored the transferability of a trained network to new molecular systems and found that finetuning with less than 10\% of the total trajectory provides a mean absolute error of the same order of magnitude as that when training from scratch on the full molecular trajectory.

\end{abstract}
\section{Introduction}

Modeling the structural dynamics of molecules is important in a number of applications, from protein folding \cite{alquraishi2019alphafold} to molecular devices \cite{balzani2006molecular} and adaptive materials \cite{lehn2015perspectives}. Molecular dynamics simulations often rely on parameterized atomistic potentials, which can be inaccurate compared to \textit{ab initio} simulation methods. However, dynamic \textit{ab initio} simulations come with high computational cost. Recently, deep learning techniques have been sought to obtain both high accuracy and low cost during inference.

Neural network potentials, which predict the molecular energy and forces on the atoms of a static structure, have become increasingly popular for use in molecular dynamics simulations \cite{behler2014representing,ko2021fourth,schutt2018schnet,smith2017ani,zhang2019embedded}. The structural dynamics are not predicted directly by such potentials, but are instead derived by  numerically solving Newton's equations of motion using the energies and forces provided by the potential. The evolution of the structure is determined from the prior step and long-term time dependencies are not considered. Therefore, it is useful to explore additional methods for dynamic structure generation that do consider such long-term dependencies.

The description of molecules as molecular graphs in which atoms translate to vertices and bonds translate to edges is a natural choice. When describing molecules in such a way, molecular structure generation becomes a graph generation problem. A number of works have explored the generation of static molecular graphs using various graph neural network (GNN) architectures. Notably, autoencoders have shown great success in the generation of chemically valid molecular graphs  \cite{bresson2019two,chang2019tiered,jin2018junction,kwon2019efficient,liu2018constrained,simonovsky2018graphvae}.
Another important task is the generation of molecular graphs with optimized properties \cite{assouel2018defactor,kearnes2016molecular,liu2021graphebm,zang2020moflow}.
Reinforcement learning, in particular, has been combined with autoencoders \cite{born2019paccmann,kearnes2019decoding}, graph convolutional networks (GCNs) \cite{khemchandani2020deepgraphmolgen,you2018graph}, or flow models \cite{luo2021graphdf,shi2020graphaf} to generate static optimized molecular graphs for a variety of applications.


Dynamic molecular graph generation with deep generative models has not been as well explored as the static molecular graph generation problem. However, dynamic GNNs have been well studied for applications such as communication and transportation networks, recommendation systems, and epidemiology \cite{skarding2020foundations}. Such networks for dynamic graphs represented in discrete time intervals often couple GNN architectures, which learn local topological characteristics of a single snapshot, with recurrent neural networks (RNNs), which capture the temporal evolution \cite{wu2020comprehensive}. For example, Zhang et al.~showed the application of this scheme to the protein folding process \cite{zhang2020disentangled}.

In this work, we take a cue from the temporal link prediction model for weighted dynamic graphs developed by Lei et al., in which a generative adversarial network (GAN) was used to predict changes in a graph's adjacency matrix \cite{lei2019gcn}. The generator was composed of a GCN and long short-term memory (LSTM) cells, while the discriminator was a fully connected feed-forward network. Their temporal link prediction model was applied to relatively sparse communications and mobility networks. In this work, we apply the same generator architecture to temporal edge weight prediction in complete molecular graphs. We explore the transferability of a trained network to new molecular systems and find that finetuning with less than 10\% of the total trajectory provides a mean squared error (MSE) of the same order of magnitude as obtained when training from scratch on the full molecular trajectory.

\begin{figure}\centering
\includegraphics[width=0.75\textwidth]{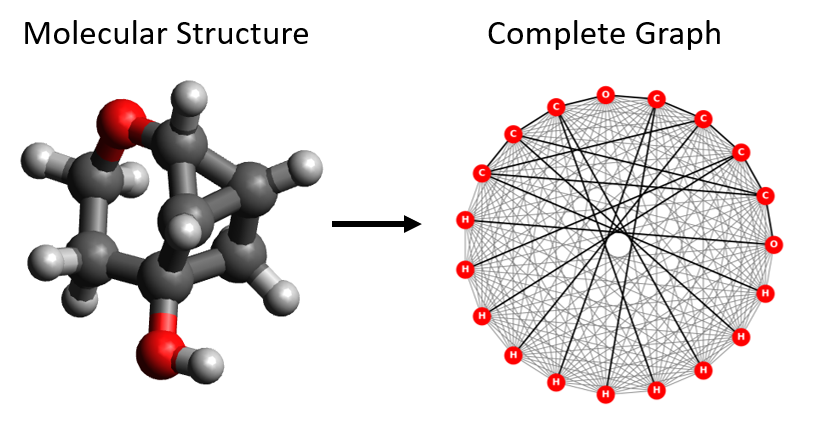}
\caption{Representation of a molecular structure as a complete graph. In the structural representation, carbon atoms are shown in dark gray, hydrogen atoms in light gray, and oxygen atoms in red. Connections between atoms represent covalent bonds. In the complete graph depiction, vertices are shown in red, edges corresponding to covalently bonded atom pairs are shown in black, and edges between non-bonded atom pairs are shown in gray.} \label{fig:molgraph}
\end{figure}

\section{Methods}
\subsection{Molecular Graph Description}
Our dynamic GNNs were trained on a subset of the ISO17 dataset \cite{ramakrishnan2014quantum,qmdatasets}. The total dataset consists of 129 molecules, all of which are constitutional isomers of \ce{C7O2H10}. In chemistry,  constitutional isomers are molecules with the same chemical formula, but differ in atom-to-atom connectivity. For each molecule, the trajectory was simulated for 5 ps with a resolution of 1 fs using Density Functional Theory (DFT) with the Perdew-Burke-Ernzerhof (PBE) functional \cite{perdew1996generalized} as implemented in the Fritz-Haber Institute \textit{ab initio} simulation package \cite{blum2009ab}. As a result, each of the 129 isomers has data containing 5,000 conformational geometries, energies, and forces. Our networks were trained on a subset of 26 isomers in which the order of the time-steps in the geometries was preserved. It is important that the order is known so that our networks can learn the time-dependent changes in molecular structure. 

\begin{figure}\centering
\includegraphics[width=0.8\textwidth]{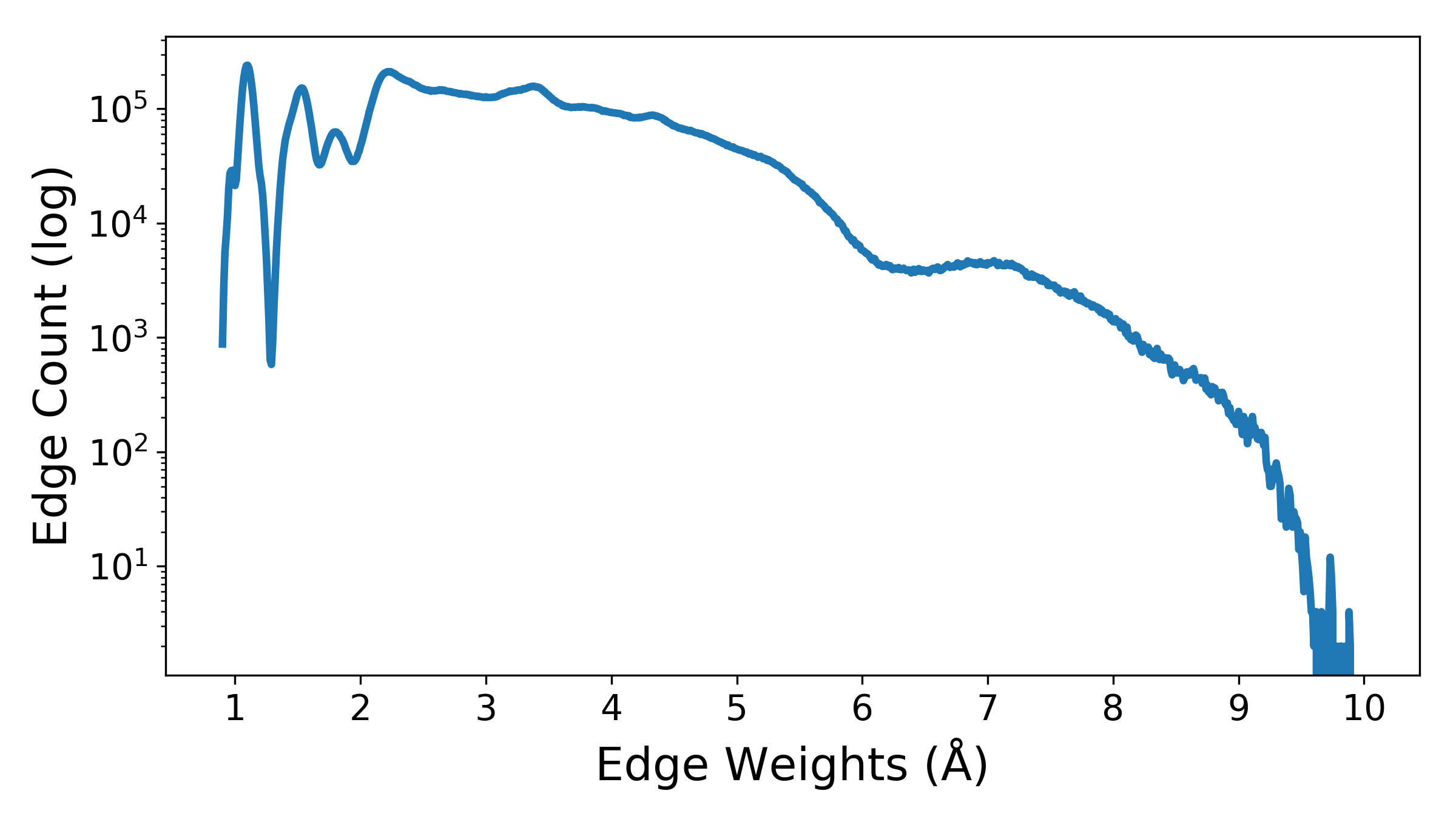}
\caption{ Distribution of edge weights in \AA~for all isomers.} \label{fig:edgeweights}
\end{figure}

In the basic graph representation of a molecule in which vertices are atoms and bonds are edges, molecular dynamics cannot be accurately described because conformational changes are not represented. 
Therefore, we define a molecule as an undirected complete graph $G=(V,E,w)$. Each vertex in the graph represents an atom, giving a total of 19 vertices per isomer graph. An example depiction of a complete molecular graph from our ISO17 subset is shown in a circular layout in Figure \ref{fig:molgraph}. Although the atom type is shown in figure, this is strictly a visual representation of an isomer molecule. Atom information is not given to our network during training. Every pair of distinct vertices $V$ is connected by a unique edge $E$, weighted by the Euclidean distance between the two connected atom vertices. In our ISO17 subset, the edge weights range from 0.839 to 9.894 \AA~(see Fig. \ref{fig:edgeweights} for the distribution of edge weights over all isomers).  
The adjacency matrix of each graph is a symmetric, dense matrix of size $N_{\mathrm{atoms}} \times N_{\mathrm{atoms}}$ --- which for the isomers in our ISO17 subset is $19 \times 19$ --- with zeros on the diagonal (no self loops).

\subsection{Network Architecture}
Our goal is to predict $\mathbf{A}_{t_{n+1}}$ from the sequence of $\mathbf{A}_{t_0}$ to $\mathbf{A}_{t_n}$. For this, we build upon the work of Lei et al.~for the temporal edge prediction of weighted dynamic networks \cite{lei2019gcn}. The graph prediction network is made up of two components: a GCN and LSTM cells. The GCN component models the topological structure of individual snapshots, while the LSTM component models the temporal structure of the full sequence.

The adjacency matrix $\mathbf{A}$ of each snapshot is first passed through the GCN. The hidden representation $\mathbf{X}$ of each GCN unit is obtained as follows:

\begin{equation}
    \mathbf{X} = f(\mathbf{\hat{D}}^{-1/2} \mathbf{\hat{A}} \mathbf{\hat{D}}^{-1/2}\mathbf{Z}\mathbf{W})
\end{equation}

\noindent where $\mathbf{\hat{D}}^{-1/2} \mathbf{\hat{A}} \mathbf{\hat{D}}^{-1/2}$ is the approximated graph convolution filter, $\mathbf{\hat{A}} = \mathbf{A} + \mathbf{I}$, $\mathbf{W}$ is the weight matrix, and $f$ is the activation function.

The hidden representation \textbf{X} of each snapshot is passed into the corresponding LSTM cell, where it travels through a forget gate $f_t$, an input gate $i_t$, an input modulation gate $\Tilde{c}_t$, and an output gate $o_t$ \cite{hochreiter1997long}:

\begin{equation}
    f_t = \sigma (\mathbf{W}_f \cdot [h_{t-1}, \mathbf{X}_t] + b_f )
\end{equation}
\begin{equation}
    i_t = \sigma (\mathbf{W}_i \cdot [h_{t-1}, \mathbf{X}_t] + b_i )
\end{equation}
\begin{equation}
    \Tilde{c}_t = \mathrm{tanh}(\mathbf{W}_c \cdot [h_{t-1}, \mathbf{X}_t] + b_c )
\end{equation}
\begin{equation}
    o_t = \sigma (\mathbf{W}_o \cdot [h_{t-1}, \mathbf{X}_t] + b_o )
\end{equation}

\noindent where $h_{t-1}$ is the hidden state at the previous time step, \textbf{W} are the weights of the corresponding gate, $b$ is the bias, and $\sigma$ is the sigmoid activation function. The current cell state $c_t$ and hidden state $h_t$ are then updated as:

\begin{equation}
    c_t = f_t \odot c_{t-1} + i_t \odot \Tilde{c}_t
\end{equation}
\begin{equation}
    h_t = o_t \odot \mathrm{tanh}(c_t)
\end{equation}

\noindent where $\odot$ denotes element-wise multiplication.

\begin{figure}\centering
\includegraphics[width=0.4\textwidth]{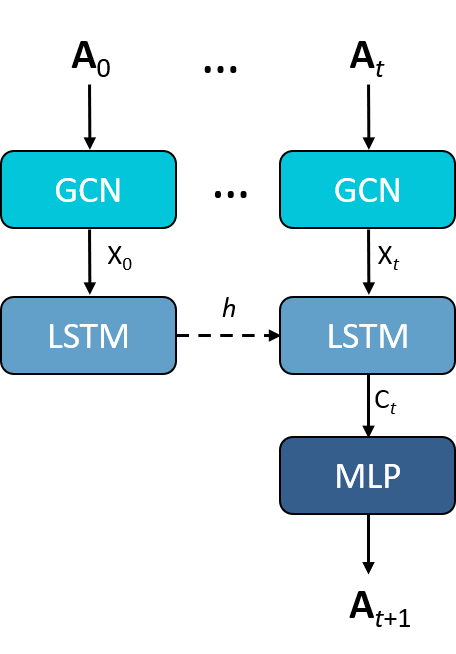}
\caption{Diagram of the dynamic graph prediction network used in this work. A sequence of adjacency matrices \textbf{A}$_{0}$ to \textbf{A}$_{t}$ are fed into a graph convolutional network (GCN). The embeddings from the GCN are passed into long short-term memory (LSTM) cells, which pass along hidden states $h$. The cell state $C_t$ is passed to a multilayer perceptron (MLP), which outputs \textbf{A}$_{t+1}$} \label{fig:network}
\end{figure}

The hidden state $h_t$ is used to update the LSTM cells during training, while the final cell state $c_t$ is fed into a multilayer perceptron (MLP), which generates $\mathbf{A}_{t_{n+1}}$. 
A diagram of the dynamic graph prediction network is shown in Figure \ref{fig:network}. Here, we use a window size of 10, GCN feature size of 64, and LSTM feature size of 128. Code for the network was written using PyTorch and can be found at \url{github.com/pnnl/mol\_dgnn}.

\section{Results and Discussion}

In this section, we apply our dynamic GNN to variations of our 26 isomer subset of the ISO17 dataset. The full trajectories were split into overlapping windows of size 10. This produced 4,990 samples per isomer. The samples were then split into training and test sets in a ratio of 0.8:0.2. All networks were trained using stochastic gradient decent with the ADAM optimizer \cite{adamoptimizer} using a learning rate of $5.0 \times 10^{-3}$. All networks were trained for 5,000 epochs. 

We performed four experiments to validate our dynamic GNN. We first trained a set of networks over the conformational geometries of each individual isomer to test how well our network could predict trajectories of a single isomer. We then performed a cross-test for each of these single isomer networks on the test sets of all other isomers to examine the generalizability. Next, we trained a single network on a training set of all isomers combined to examine the ability of the network to predict the evolution of multiple isomers. Finally, we performed a finetuning study, where we trained the network for a limited number of epochs on a small subset of samples from the training sets of individual isomers starting from the weights of the single-isomer networks of opposing isomers. 
The single-isomer, cross-test, and finetuned networks were trained with a training set of 4,000 samples per isomer, a test set of 980 samples per isomer, and a batch size of 400. The all-isomer network used a training set of 103,792 samples, a test set of 25,948 samples, and a batch size of 499. 

Our first set of evaluation metrics is based on the predicted edge weights between atom pairs. Errors are presented as the mean squared error (MSE), mean absolute error (MAE), and percent error (PE) in the edge weights. MSE and MAE are benchmark error representations in many machine learning applications, as well as in the field of chemistry. For molecular structure computations, an MAE in atom-to-atom distances on the order of 0.010--0.020 \AA~is considered ``chemically accurate''. Though MSE and MAE allow comparison between different network architectures and training structures, these metrics assume the target absolute errors should be similar for all predicted outputs. In the case of molecules, there exists nuances within the weight distributions. For example, the distances between covalently bound atom pairs exist in a much narrower range than the distances between non-bonded atom pairs. It is important to quantify the relative difference, as small errors in the weights of covalently bound pairs has a much larger effect on the molecular energy than the same size errors in non-bonded pairs. We therefore calculate the PE, defined as 
\begin{equation}
    \mathrm{PE}=100 \times \frac{|w_t - w_p|}{w_t}
\end{equation}

\noindent where $w_t$ is the true edge weight and $w_p$ is the corresponding edge weight predicted by our network. PE is further distinguished for bonded and non-bonded atom pairs, as we observed a difference in the errors between the two. 

Finally, a graph similarity metric is used to validate the predicted graphs as a whole. We use the eigenvalue method described by Koutra et al.~\cite{koutra2011similarity} to quantify the similarity of the generated graph to the true graph at that step in the trajectory. To calculate the eigenvalue similarity (S), we compute the eigenvalues ($\lambda$) of the combinatorial laplacians of the two graphs $G_1$ and $G_2$ under study. S of $G_1$ and $G_2$ is then computed as
\begin{equation}
    \mathrm{S}=\sum_{i=1}^{k}(\lambda_{1i}-\lambda_{2i})^2
\end{equation}
over the top \textit{k} eigenvalues that contain 90\% of the energy. This similarity metric is unbounded, [0, $\infty$). Isomorphic graphs show S=0, and S increases with increasing dissimilarity.

\begin{table}\centering
\caption{Test set mean squared error (MSE), mean absolute error (MAE), and percent error (PE) of edge weights and the eigenvalue similarity (S) of the molecular graphs produced in the four experiments described in this work.}\label{tab:results}
\setlength{\tabcolsep}{10pt}
\begin{tabular}{|l|c|c|c|c|}
\hline
Experiment &  MSE & MAE & PE &  S \\
\hline
Single isomer & 0.00029 & 0.017 & ~3.2\%  &  0.108 \\
Cross-set     & 0.01334 & 0.115 & 30.6\%  &  8.308 \\
All isomers   & 0.00687 & 0.083 & 19.7\%  &  5.099 \\
Finetuned     & 0.00067 & 0.023 & ~4.6\%  &  0.152 \\
\hline
\end{tabular}
\end{table}

\begin{figure}\centering
\includegraphics[width=\textwidth]{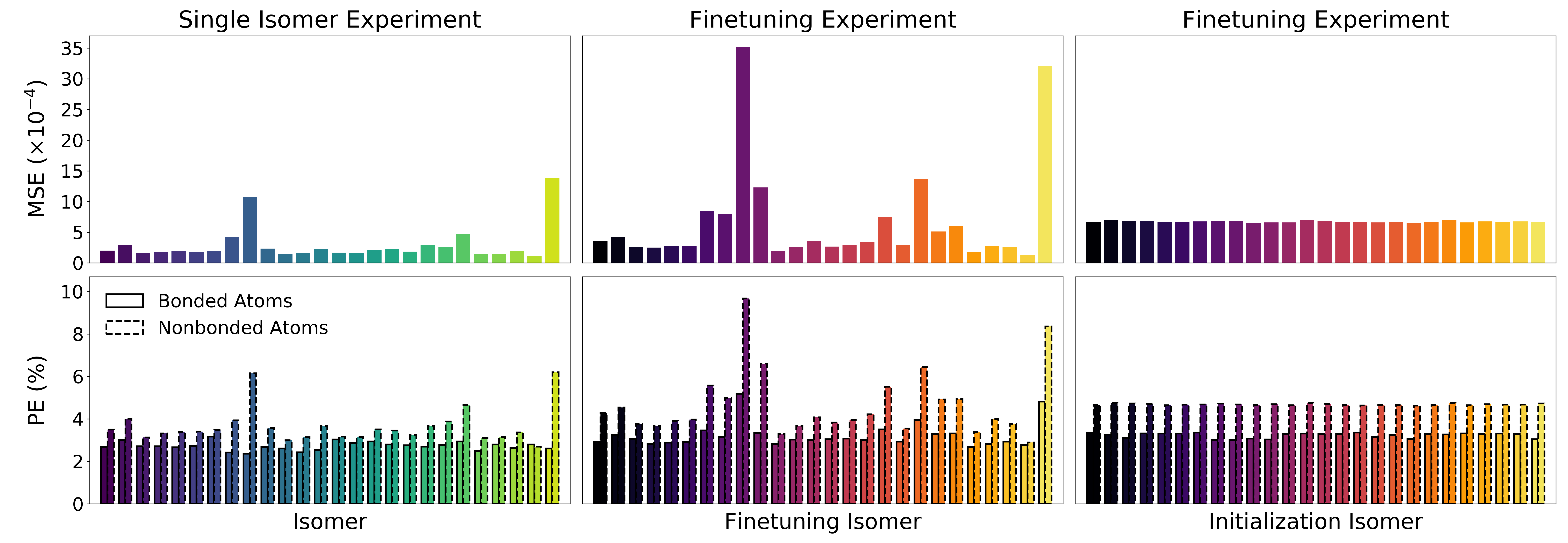}
\caption{The mean squared error (MSE) and percent error (PE) of bonded (solid bars) and non-bonded (dashed  bars) atom pairs for each test set isomer. The left-most plots show the test set results of training on 80\% of the full trajectory of a single isomer. The plots in the middle and on the right show the results of finetuning those networks on a 10\% subset of a different isomer. The middle plots show the errors aggregated by the isomer set used to finetune the network, while the right-most plots show the errors aggregated by the initialization isomer used to train the base network before finetuning.} \label{fig:msepe}
\end{figure}

We first trained separate dynamic GNNs for each individual isomer to examine the prediction accuracy of our network. The results from inference on the test set are shown on Table \ref{tab:results}. Over all isomers, the networks achieved an average MSE of $2.9 \times 10^{-4}$ \AA$^2$, MAE of $1.7 \times 10^{-2}$ \AA, and PE of 3.2\%.  We further separated PE into two categories: bonded and non-bonded atom pairs. For each isomer, the MSE and PE on the test set are shown in Figure \ref{fig:msepe}. We observed that non-bonded atoms have a significantly higher PE than bonded atoms. This may be because bonded atom distances are typically concentrated within a small range of ~1--3 \AA, while the more numerous non-bonded atom pairs have a much larger distribution between 3--10 \AA. We also examined the similarity S of the graphs resulting from the generated adjacency matrices and observed an average S of 0.108, indicating high similarity between the predicted and true complete molecular graphs.

The MAE of the single isomer networks is within the range of ``chemical accuracy'' for computed molecular structures. Therefore, we next tested the generalizability of our network by using the trained networks to infer on the test sets of the other isomers. The average over all the networks of these cross-test results are displayed in Table \ref{tab:results}. We observed a cross-test MSE of $1.334 \times 10^{-2}$ \AA$^2$, MAE of $1.15 \times 10^{-1}$ \AA, PE of 30.6\%, and S of 8.308. This poor generalizability indicates that our dynamic GNN trained on a single isomer alone is not suitable for predicting trajectories of other isomers.

In an attempt to improve the generalizability, we combined the training sets of all isomers. Training on this large, multi-isomer dataset improved the associated error metrics, as shown in Table \ref{tab:results}, but still gave an unsatisfactory PE of 19.7\% and S of 5.099.

With the ISO17 dataset, we have the luxury of training individual networks over the trajectories of each isomer. However, as the size of the data increases, so does training time and computational cost. In addition, collection of the trajectories by DFT is non-trivial and also has high associated computational costs. It is therefore inefficient to train a network from scratch on a large number of snapshots to predict the subsequent trajectory of new molecules. Our proposed solution is to finetune our trained networks for a reduced number of epochs on a small number of new snapshots. We used the weights from the networks trained on the individual isomers to initialize training on a new isomer. By initializing the weights from a pre-trained network, we observed improved performance using a much smaller training set for the new isomer. The networks were finetuned on a training set of 400 samples as compared to the 4,000 samples used to train the single isomer networks from scratch. Each isomer model was finetuned across all other isomers for a total of 500 epochs -- an order of magnitude fewer epochs than used to train the single isomer networks. The average MSE, MAE, PE, and S for each isomer are given in Table \ref{tab:results}. Figure \ref{fig:msepe} shows the MSE and PE of bonded and non-bonded atom pairs grouped by the finetuning isomer, as well as by the initialization isomer. Finetuning showed an average MSE of $6.7 \times 10^{-4}$ \AA$^2$, MAE of $2.3 \times 10^{-2}$ \AA, PE of 4.6\%, and S of 0.152. These error metrics are of the same order of magnitude as those of the networks trained and tested on a single isomer, which indicates that finetuning provides comparable predictions while training on a much smaller set of data for a reduced number of epochs. Notably, Figure \ref{fig:msepe} shows a uniform distribution for both MSE and PE over all initialization isomers. Therefore, finetuning is not dependent on which initial isomer is selected for pre-training.

\section{Conclusions}

We have proposed the use of a dynamic GNN to predict time-dependent changes in molecular structure. Networks were trained on the trajectories of a subset of isomers from the ISO17 dataset. Trajectories are predicted by learning atom-to-atom distances from the complete graph representation of an isomer's molecular structure. 

We first trained networks on each individual isomer. Our trained networks achieved an average MSE of $2.9 \times 10^{-4}$ \AA$^2$, MAE of $1.7 \times 10^{-2}$ \AA, and PE of 3.2\%. Non-bonding PEs are significantly higher than bonding PEs, hypothesized to be caused by the larger range and sample size of non-bonded edges. 

Multiple networks were trained to test the transferability and generalizability of our dynamic GNN. Cross-set networks were evaluated, where a trained network from one isomer was used to predict trajectories of a different isomer. Cross-set networks achieved an average MSE of $1.334 \times 10^{-2}$ \AA$^2$, MAE of $1.15 \times 10^{-1}$ \AA, and PE of 30.6\%. These results suggest our GNN trained on one isomer alone is not suitable to predict trajectories of other isomers. We tried improving generalizability by training a network over a large training set that included all isomers. This all-isomers network achieved an average MSE of $6.87 \times 10^{-3}$ \AA$^2$, MAE of $8.3 \times 10^{-2}$ \AA, and PE of 19.7\%. This large-scale training still proved to be unsatisfactory. Finally, we finetuned the single-isomer networks on a small subset of samples of different isomers. These networks achieved an average MSE of $6.7 \times 10^{-4}$ \AA$^2$, MAE of $2.3 \times 10^{-2}$ \AA, and PE of 4.6\%. Our finetuned networks performed similar to the single isomer networks, suggesting that finetuning is a viable method for predicting trajectories of new isomers. Furthermore, we showed that the results are independent of which isomer is chosen for pre-training. 

Traditionally, accurate molecular trajectories are calculated at high computational cost. We propose using dynamic GNNs as a method to predict time-dependent trajectories of molecules. Our networks achieved accurate, time efficient, and cost efficient predictions within the realm of ``chemical accuracy''. For these reasons, we expect dynamic GNNs to play a role in the future of molecular dynamics.

\bibliographystyle{splncs04}
\bibliography{refs}

\end{document}